\documentclass[applsci,article,accept,pdftex,moreauthors]{Definitions/mdpi} 

\firstpage{1} 
\pubvolume{0}
\issuenum{0}
\articlenumber{0}
\pubyear{2022}
\copyrightyear{2022}
\datereceived{} 
\dateaccepted{} 
\datepublished{} 
\hreflink{https://doi.org/} 

\Title{Multimodal Classification of Teaching Activities from University Lecture Recordings}
\TitleCitation{Multimodal Classification of Teaching Activities from University Lecture Recordings}

\Author{Oscar Sapena $^{\dagger}$\orcidA{} and Eva Onaindia *$^{,\dagger}$\orcidB{}}

\AuthorNames{Oscar Sapena and Eva Onaindia}

\AuthorCitation{Sapena, O.; Onaindia, E.}

\address [1]{Valencian Research Artificial Intelligence Institute, Universitat Politècnica de València, Camino de Vera s/n, 46022~Valencia, Spain; osapena@dsic.upv.es}

\corres{\hangafter=1 \hangindent=1.05em \hspace{-0.82em}Correspondence: onaindia@dsic.upv.es}

\firstnote{\hangafter=1 \hangindent=1.05em \hspace{-0.82em}These authors contributed equally to this work.}

\abstract{The way of understanding online higher education has greatly changed due to the worldwide pandemic situation. Teaching is undertaken remotely, and the faculty incorporate lecture audio recordings as part of the teaching material. This new online teaching--learning setting has largely impacted university classes. While online teaching technology that enriches virtual classrooms has been abundant over the past two years, the same has not occurred in supporting students during online learning. {To overcome this limitation, our aim is to work toward enabling students to easily access the piece of the lesson recording in which the teacher explains a theoretical concept, solves an exercise, or comments on organizational issues of the course. To that end, we present a multimodal classification algorithm that identifies the type of activity that is being carried out at any time of the lesson by using a transformer-based language model that exploits features from the audio file and from the automated lecture transcription. The experimental results will show that some academic activities are more easily identifiable with the audio signal while resorting to the text transcription is needed to identify others. All in all, our contribution aims to recognize the academic activities of a teacher during a lesson.}}


\begin{document}

\section{Introduction}

Within the context of education, and~more specifically in higher education, the~global pandemic situation over the past two years has led to a widespread use of remote teaching/learning and lecture recordings as part of the teaching material. This new online teaching--learning setting has largely impacted university~classes.

Regardless of the type of educational system prevailing in each country, university classes typically follow a lecture-based instructional approach where the lecture is verbally delivered by an instructor who supports their academic discourse by using a slide presentation and/or a writing surface. This has not changed much over the pandemic, with the most notable exception that lectures are now recorded and so students replace note-taking by video-watching when~studying.

While online teaching technology tools have highly improved over the past two years, the~same has not occurred in supporting students during online learning. Lecture recordings have become a key learning means for millions of students regardless of the availability of in-person class attendance. Everyone wishes to have access to a backup material that one can resort to for post-lecture~learning. 

The benefits of lesson recordings to support the learning experience of students are diverse such as providing content that can be reviewed multiple times, accessibility to material in case of an impossibility to attend in-person or~focusing on listening to the lecturer rather than taking notes. Despite these clear benefits, several investigations have revealed that using learning technology is one of the most common challenges that students face during online learning~\cite{Rasheed20, BarrotLR21}.

The impact of lecture videos on students' academic performance has been investigated from different perspectives such as analyzing the actual time spent on the usage of recorded lectures in relation to lecture attendance and the effect on exam performance, or~studying the impact of combining lecture videos with written materials on students' outcomes~\cite{LeadbeaterSCN13,Bos16, Robertson20}. There have also been works that explore if the usage of video recordings by the students varies for different subjects or if it is different for subgroups of students~\cite{Sarsfield_Conway_2018}. However, finding literature on how to support students to engage with the recordings is less frequent, mostly because students have different learning styles, different study strategies and skills. There is though one functionality that would be very helpful for post-lecture learning from class recordings (or any recording in general) and that is the ability to find and view the desired contents in the recorded lecture video. This becomes particularly relevant when the recording is used to strengthen the contents acquired during an in-person session. Moreover, providing class recordings for supplementary use has been lately advised in regular nonpandemic educational environments, since no negative effect of the use of recordings has generally been evidenced in university students~\cite{Nordmann19}.

The work presented in this paper intends to be a first step toward facilitating the view of particular contents in a class recording without the need to play the video back and forth until the desired segment is found. Imagine, for~instance, a~student who wants to find in a two-hour video recording the segment where the lecturer delivers a task assignment or a student who just wants to watch the part of the video in which the lecturer is solving an exercise. In~general, replaying the audio-video recording repeatedly is not an effective studying method for university students, who need to adjust their study strategies to suit this~mode.

Our contribution puts the focus on a classification model that identifies the type of teaching activity (solving an exercise, explaining a theoretical concept, talking about a task assignment, interacting with students, etc.) that is being undertaken by the lecturer at each instant of the video recording. We believe that classifying the type of academic activity of the lecturer is a first crucial phase towards the development of a tool to facilitate students finding specific content in online recordings. To~that end, we propose a novel multimodal classification model that identifies segments of the class recording by jointly exploiting the audio signal and the automated transcription of the recorded~lecture.

This paper is structured as follows. The~next section presents a literature review of approaches that classify the spoken discourse from a recording using audio and text features. Section~\ref{materials} presents the materials used for the design of the multimodal classification model including the proposed classification of teaching activities and~a description of the audio and transcription files of the recordings. Section~\ref{methods} is devoted to describing the methods used for the feature extraction as well as the architecture of the classifier. Section~\ref{results} presents the results of the experimental evaluation. Section~\ref{discussion} highlights the main results of our approach in relation to previous studies, and~Section~\ref{conclusions} concludes and outlines future research~lines.



\section{Literature~Review}
\label{literature_review}

This section outlines the main approaches that use the audio signal and automated transcriptions of a recorded speech for different educational purposes. Most of the approaches exploit machine learning models, specifically deep learning (DL) techniques, which are currently being extensively applied in a large variety of educational tasks such as predicting student academic performance~\cite{Balaji21} or assessing the performance of educational institutions~\cite{Alam21}.

We divide this section in three parts: the first one is devoted to approaches that only exploit the audio signal of the recordings, the second one to approaches that use textual analysis techniques in education, and~the third one to hybrid approaches that attempt a combination of both data sources, audio and~text.

\subsection{Audio-Based~Classification}
\label{audio_classification}

Educational tools for analyzing the classroom academic discourse are scarce and~the few existing ones mostly use the audio signal of the recordings with the aim to distinguish the teacher lecturing from the student participation in class. 
LENA (Language ENvironment Analysis) is a system that records and~analyzes classroom discourse to provide teachers a timely feedback that improves their skills in classroom discourse management~\cite{WangPMC14,LENAOrg14}. LENA is particularly oriented to child language development. Teachers using LENA wear a proprietary wearable audio recorder while teaching a regular math lesson to small children to collect speech data. LENA implements a speech recognition system aimed to identify three common discourse activities: teacher lecturing, whole class discussion and student group work~\cite{WangPMC14}, and~it has primarily been used to assess small children's language environment~\cite{Cristia21}. No transcription of the words in the recordings is provided by LENA, instead it produces a diarization of who is talking and when, according to the predetermined categories~\cite{GANEK201877}.

Another interesting audio-based classification system is the project Decibel Analysis for Research in Teaching (DART) that analyzes the volume and variance of audio recordings of science technology engineering mathematics (STEM) courses to predict how much time is spent  on single-voice (e.g., lecture), multiple-voice (e.g., pair discussion), and~no-voice (e.g., clicker question thinking) activities~\cite{Owens3085}. DART aims to identify the types of activities that are going on in a classroom based exclusively on sound~waveforms. 

Audio-based classification has also been used for assessing time devoted to lecturing and student discussion, specifically in a flipped classroom setting~\cite{SuDWLM19}. Similar to DART, in~the latter cited work, the authors use multiple audio recorders to detect segments of lecture that are primarily the lecturer's speech, while segments of discussion comprise students' speech, silence and~noise. 

All the aforementioned systems apply speech recognition and methods from the field of audio segmentation with the aim of analyzing signal intensity (volume) and variations in the sound of a classroom. This analysis is used to identify the speaker or classify the sound into categories accordingly to detected voices. This way, a~single voice is associated to teacher talk, multiple voices to students talk, no voice to silence, and~other indistinguishable voices to murmuring or overlapping speech. Additionally, both LENA and DART are supervised learning systems that require human annotations of the recorded speech. LENA uses a random forest algorithm while DART uses support vector machine~techniques.

\subsection{Text Analysis in~Education}
\label{text_analysis}

Audio segmentation and classification are helpful to distinguish the lecturer's speech from group discussion or teamwork, but~they do not provide significant information for recognizing teaching activities in classes that follow a type of lecture-based learning approach. The~academic lecture is mostly considered an expository genre where interaction with students is less frequent than in other classroom genres like seminars, tutorials or oral presentations~\cite{Fortanet05b}; yet, it is becoming more and more relevant due to the increasing internationalization of higher education~\cite{Fortanet05a} and its simplicity to be adapted to an online~format. 

Audio transcription and text classification become increasingly relevant for  \linebreak analyzing academic discourse. Nowadays there exists a wide range of software that \linebreak  automatically transcribes audio and video using high-end AI engines. It is even possible to find transcription tools for particular contexts such as medical transcription or \linebreak  supporting sales teams. When it comes down to education, many universities offer their own automated transcription services ({some examples of university services of automated transcriptions include:  \url{https://guides.nyu.edu/QDA/transcription}; \url{https://www.nottingham.ac.uk/dts/researcher/applications-and-tools/automated-transcription.aspx};$\;$    \url{https://www.bentley.edu/centers/user-experience-center/transcription-tools-qualitative-data-uxr};   \url{https://www.universitytranscriptions.co.uk/}}) (accessed 18 April 2022), which usually provide transcriptions suited to the technical, scientific, or~social language used in the delivery of~classes.

Let us now turn our attention to the analysis of text coming from transcriptions of university lecture recordings. Linguistics acknowledges that topic-oriented university lectures are categorized by features that capture the informational purpose of the speech (theoretical information and examples of practical application) and features that display the spoken discourse such as rhythm, intonation, speed of utterance, pausing and \linebreak  phrasing~\cite{Young1995,Crystal95,Csomay00}. More recent studies show that university language in academic speech includes vocabulary patterns, the~use of lexical-grammatical syntactic features, discourse connectors or lexical bundles and formal or informal language when required~\cite{Biber06,Malavska16}.

Textual analysis techniques in education have been successfully applied to analyze students' answers and make better judgment on their performance~\cite{Cunningham18}, to extract interesting and high-quality information from unstructured text~\cite{Ferreira19} and mainly to topic modeling for different purposes, such as discovering important themes and patterns for formative assessment of students’ learning~\cite{Chen16}, analyzing teachers' understanding~\cite{Chen16} or retrieving relevant educational material~\cite{Wang15}. 

All the existing approaches for textual analysis employ natural language processing (NLP) techniques. Recently, the~use of transformer-based models has gained great popularity.  The~revolution of deep learning has produced a dramatic effect in NLP thanks to novel end-to-end architectures that do not require any prior knowledge on language nor the use of traditional language processing tasks such as tokenization, syntactic parsing, stemming, part-of-speech tagging, etc.~\cite{YoungHPC18,DeanPY18}. New architectural models such as the bidirectional encoder representations from transformers (BERT) family~\cite{Devlin18} and the generative pretrained transformers (GPT-2 and GPT-3) \cite{Radford19} are among the most popular models. These two models, and~others recently proposed, are based on the transformer architecture, a~DL model that adopts the self-attention mechanism whereby different importance to each piece of the input data is given accordingly to its significance~\cite{VaswaniSPUJGKP17}. Incorporating the attention mechanism in network models has generated significant improvements in data augmentation and text classification~\cite{Miao22}.

The great advantage of using transformer models in NLP lies in an efficient computation of sequence-to-sequence tasks while facilitating the handling of long-range~dependencies.

\subsection{Multimodal Classification of~Conversations}
\label{multimodal}

There are hardly any investigations that jointly explore the audio signals and the text of automated transcriptions in the context of education. Most research in multimodal approaches are oriented toward estimating the speaker's emotion in an audio conversation~\cite{BHASKAR2015635,Yoon18}, using the sound and spoken content of an emotional dialogue to obtain a better understanding of speech data. Since the focus is on classifying the emotional content of speech, these approaches typically work with a fixed vocabulary of words that identify an emotional category (e.g., ``happy'', ``sad'', ``angry''). Therefore, the~scope of the content analysis achievable with these tools is significantly more limited than with the techniques exposed in Section~\ref{text_analysis}, as the language used in these applications is delimited to particular words with an emotional~burden.

Textual-acoustic feature representation has also been applied to sentence-level speech classification for detecting intention in the speech of a medical setting~\cite{GuChen17}, and~to music genre classification as in the work presented in~\cite{OramasBNS18}, where authors showed that the learning of a multimodal feature space increased the performance of pure audio~representations.

It is more common, however, to~find multimodal educational systems that use human rather than automated transcriptions. Some studies have evaluated the performance of the LENA system (see Section~\ref{audio_classification}) applied to native French-speaking young children using audio recordings and their manually transcribed files~\cite{Canault16b}. For~systems devoted to the language development of small children (children from birth to 3 years as in the case of LENA), it is affordable to have a professional team to produce accurate and reliable transcription of the audio recording files~\cite{gilkerson2008}.

Broadly speaking, we can conclude that multimodal classification is based on the audio signal as the primary data source and it is complemented by the extraction of text features. Furthermore, we can also affirm that this type of classification has mainly been explored in particular contexts of application (emotional, medical, music) that feature their own professional or technical~jargon.

\subsection{State of the Art in Our~Approach}

In this section, we briefly emphasize the results from the literature review our approach relies on and how our contribution advances the state of the~art.

Similar to the approaches referenced in Section~\ref{multimodal}, we also propose a classification model that builds on the audio and text of a lecture recording. Even though our system is designed for the classification of teaching activities in higher education, it is intended for a broad applicability across a variety of different university subjects such as mathematics, oceanographic physics or electronic devices. Hence, unlike some of the reviewed approaches, we are not confined to some particularly specific~language.

We observe that our classification approach uses automated transcriptions, not manually transcribed text. This introduces a higher degree of difficulty due to the general lack of accuracy of automated transcription vs. human transcriptions but, on~the other hand, it creates a widespread tool, as it can be used with any automated transcription~service.

Finally, we highlight that our aim is to recognize teaching activities across courses of different nature delivered in a technical university, thus our focus is not on topic modeling but on discourse analysis, that is, on~analyzing the used vocabulary, the~grammar or the way that sentences are constructed and the structure of the text that creates the~narrative.

\section{Materials}
\label{materials}

This section is structured in two parts: Section \ref{academic_disourse} presents the classification of teaching activities that we use in this work; then, Section \ref{problem_definition} details the audio and transcription files of the class~recordings.

\subsection{Activities in Spoken Academic~Lecture}
\label{academic_disourse}

In this section, we introduce a classification of teaching activities identifiable from academic spoken discourse in university classes. Specifically, our interest is to come up with a set of academic labels which characterize teaching STEM subjects and which are useful for students to be able to access the contents of the syllabus as well as to easily find any organizational issue related to the~course.

Our proposal is inspired by the typical structure of a lecture presented by Malavska in~\cite{Malavska16} and the academic labels used by Diosdado et al. in~\cite{DiosdadoRO21}. Our set of labels is organized according to whether the label identification is more dependent on the audio signal or on the automated~transcription.

\begin{figure}[H]
    \centerline{\includegraphics[scale=0.5]{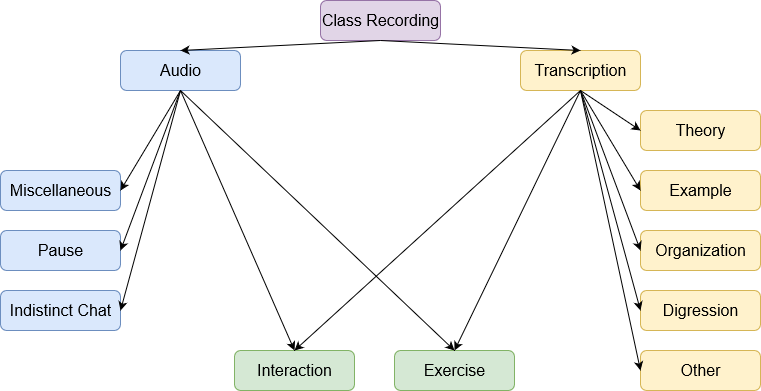}}
    \caption{Hierarchy of academic~labels.}
    \label{fig:LabelHierarchy}
\end{figure}

Figure~\ref{fig:LabelHierarchy} shows the hierarchy of labels. The~labels under the 'Audio' category are used to filter out sounds from the audio file which do not feature voices or when the recorded voices are murmurs, which are meaningless for our task. This includes sections of the audio file resulting from a muted microphone or microphone feedback (Miscellaneous), background noise (Indistinct Chat) or periods of silence between segments of speech (Pause).


The right branch of the tree in Figure~\ref{fig:LabelHierarchy} comprises the activities that come up during a regular expository class around the syllabus of a subject. The~activities classified only under the category 'Transcription' denote the nature and communicative purpose of the teacher's speech. Under~this category we gather the activities that typically involve an extended speech of the teacher with no interactions from the students: exposition of the theory (Theory) and illustration of theoretical concepts through concrete examples (Example), information about organizational issues, grading policy, assignments, scheduling, housekeeping, etc. (Organization), a~shift of the lecturer speech to a more personal discourse or course-related asides (Digression), or~a speech around non-course-related matters (Other).

We identified a third group of activities which were not distinguishable by analyzing separately the audio file and the transcription. Typically, these activities involved an exchange of communication between the teacher and students. We placed under this mixed category the label ``Interaction'', which represents teacher--student conversations that come up during the class, and~the label ``Exercise'', which accounts for a common activity in scientific/technical courses {(in nontechnical contexts this label could be replaced by ``Practical Activity'')}. Our experience from the visualization of multiples class recordings is that student engagement is far more frequent during problem and exercise solving than during theory exposition. This is the reason why ``Exercise'' is an activity classified under both Audio and~Transcription categories.

\subsection{Audio and Transcriptions~Files}
\label{problem_definition}

Our goal was to recognize the teaching activity that a lecturer was doing at any given time during a class. To~this end, we needed to segment the class recording (audio and transcriptions) and classify each segment in its corresponding activity. A~segment represented a  linguistic meaningful unit such as a word, a~sentence, a~paragraph or~any information unit depending on the task of the text~analysis.

We worked with recordings from university lectures delivered in Spanish which were obtained from a repository at our university  (UPV). The~classes were recorded using a camera that focused on the scaffold and the blackboard, and~a lapel microphone worn by the lecturer. With~this setup, we obtained a video and audio recording of the lecturer. The~lapel microphone captured the lecturer's voice with good quality. However, due to the characteristics of this kind of microphone, it was not possible to obtain a reliable capture of the students'~voices. 

Regarding the automatic transcription of lecture notes, we used the MLLP transcription and translation platform for automated and assisted multilingual media subtitling that provides support for the transcription of video, audio and content of the courses \linebreak (\url{https://ttp.mllp.upv.es/index.php?page=faq}) (accessed: 20 January 2022) \cite{Martinez-VillarongaAAJ13,MiroSCTJ15} .

The final dataset consisted of 34 audio files and automated transcriptions, each corresponding to a delivered class, which amounted to a total of 3773 min. 
We selected recordings from five male professors and five female professors to ensure gender variety, and~chose a  wide range of subjects, such as mathematics, oceanographic physics, digital signal processing, etc., to~ensure subject diversity. A~breakdown of the dataset by subject and gender can be found in Table~\ref{tab:videos_courses}. We manually labeled the automated transcriptions following the label hierarchy shown in Figure~\ref{fig:LabelHierarchy}.

\begin{table}[H] 
\caption{Breakdown of the dataset by minutes per course and per~gender.\label{tab:videos_courses}}
\newcolumntype{C}{>{\centering\arraybackslash}X}
\begin{tabularx}{\textwidth}{CCC}
\toprule
Course Name                & Male	                    &Female\\
\midrule
Electronic devices                  & -                                 & 330 min  \\ 
Digital signal processing            & -                                 & 360 min \\ 
Mathematics                         & 330 min                              & 344 min  \\ 
Measurement systems                 & -                                 & 120 \\
Microprocessed systems              & 360 min                             & -  \\ 
Networks and teledetection          & 564 min                              & -  \\ 
Oceanographic physics               & 600 min                              & -  \\ 
Physics                             & 90 min                               & - \\
Statistics                          & 225 min                              & 450 min  \\ 
\midrule
Total                      & 2169 min                   & 1604 min \\
\bottomrule 
\end{tabularx}
\end{table}
\unskip

\vspace{0.2cm}

We put the primary focus on the contents of the lecture, i.e.,~on the automated transcription. Our aim was to exploit powerful pretrained language models so that we could differentiate teaching activities based on a specific vocabulary, the~use of dates and the verbal form employed by the teacher. The~reason why we used an online transcription and translation platform of a research group from our own university UPV was because this system performed better in the academic speech setting, specifically because it transcribed scientific and technical expressions as well as mathematical formulae more accurately than other transcription systems we tried, such as, for~instance, the~transcription tools of YouTube or Microsoft~Teams.

Even so, the~automated transcriptions featured unwanted characteristics such as lack of punctuation, the absence of capital letters and minor errors. All these issues made our task more complex because of the noise introduced in the transcription and the absence of markers that split the transcribed text in smaller units such as, for~instance, sentences. Some transcription files, however, were manually revised and~thus have punctuation symbols, capital letters, etc. Whenever it was possible, we used the manually revised transcription due to their higher~quality.

Regarding the audio signals of the recordings, they provided key features of the lecturer's speech, such as the tone, the~cadence or the pauses between utterances. In~the following, we show the raw audio waveforms and the corresponding transcription of some of the teaching activities identified in the class recordings. The~waveforms were obtained with the Audacity tool~\cite{audacity}, a~free open-source audio editor and recorder. We examined the raw waveforms of various academic activities and analyzed their~differences.

\begin{figure}[H]
   \includegraphics[scale=0.35]{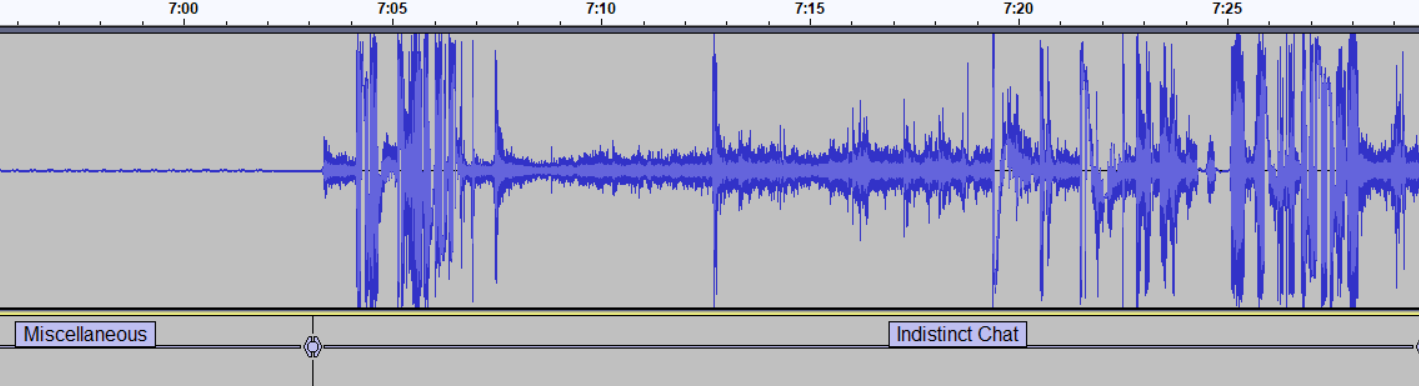}
    \caption{Audio sample of a Miscellaneous segment followed by Indistinct~Chat.}
    \label{fig:Audio_Miscellaneous+Ind_Chat}
\end{figure}

\begin{figure}[H]
    \includegraphics[scale=0.5]{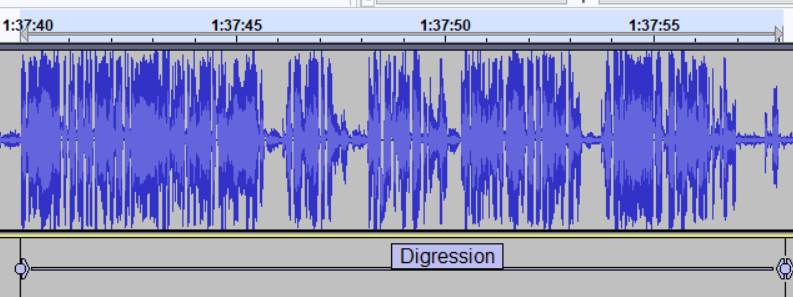}
    \caption{Audio sample of a Digression~segment.}
    \label{fig:Audio_Digression}
\end{figure}

\begin{figure}[H]
   \includegraphics[scale=0.5]{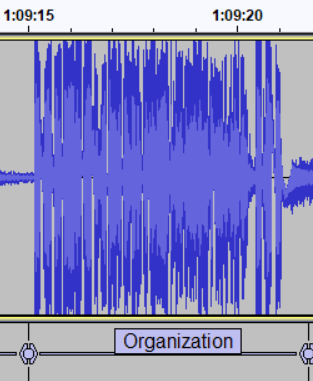}
    \caption{Audio sample of an Organization~segment.}
    \label{fig:Audio_Organization}
\end{figure}

Miscellaneous/Pause/Indistinct Chat: As we can see in Figure~\ref{fig:Audio_Miscellaneous+Ind_Chat}, a~Miscellaneous audio segment is identified by the lack of audio signal at the beginning of the recording. Since recordings are scheduled in advance, the~scheduled starting time is usually some minutes before the lecture actually begins, and~the end of the class may also be a few minutes before the scheduled ending time. Consequently, the~recordings contain several minutes where the microphone is off, leaving the recording with muted segments that we defined as Miscellaneous. We defined as a Pause a segment of audio where the lecturer does not speak for more than 2 s. However, during~this period of silence, students chat among themselves and, sometimes, they speak loud enough to be captured by the lapel microphone worn by the lecturer. We defined this occurrence as Indistinct Chat. As~can be observed in Figure~\ref{fig:Audio_Miscellaneous+Ind_Chat}, Indistinct Chat is clearly distinguishable from segments where the lecturer speaks, as~it happens in Figure~\ref{fig:Audio_Digression}, in~which the teacher makes a digression commenting on some question of a test (see transcription in Table~\ref{tab:transcription_of_Digression}), and~in Figure~\ref{fig:Audio_Organization}, in~which the words of the teacher are concerned with the organization of the class, particularly, the teacher is announcing a five-minute break (see transcription in Table~\ref{tab:transcription_of_Organization}). One can notice the distinction between Indistinct Chat and Digression or Organization by comparing the difference in the amplitude of the corresponding~waveforms.

\unskip

\unskip

\begin{table}[H]
 \caption{Transcription of Figure~\ref{fig:Audio_Digression}.}
    \label{tab:transcription_of_Digression}
        \begin{tabular}{|p{13.35cm}|}    
     \noalign{\hrule height 1pt}
        Context: The lecturer makes a comment about a detail of a test\\
       \noalign{\hrule height 0.5pt}
         Some of you have told me that I had, I put a minus on the test because I got a minus. It can't generate power, OK? It has to dissipate power. In~the tables, it will always come up as dissipated power. Right? If you get a minus, you got one of the minus signs wrong.~OK? \\
   \noalign{\hrule height 1pt}
    \end{tabular}   
\end{table}
\unskip

\unskip

\begin{table}[H]
  \caption{Transcription of Figure~\ref{fig:Audio_Organization}.}
    \label{tab:transcription_of_Organization}
    \begin{tabular}{|p{13.35cm}|}
        \noalign{\hrule height 1pt}
        Context: The lecturer tells the students to take a break and the class will continue in five minutes\\
         \noalign{\hrule height 0.5pt}
        A quarter past I want you here. Five minutes break and we start with the zener, a~quarter~past. \\
         \noalign{\hrule height 1pt}
    \end{tabular}  
\end{table}

\begin{figure}[H]
\includegraphics[scale=0.42]{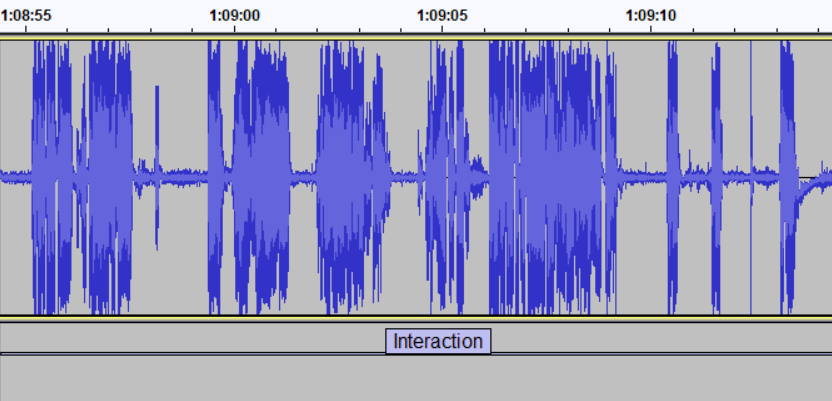}
    \caption{Audio sample of an Interaction~segment.}
    \label{fig:Audio_Interaction}
\end{figure}

Interaction: In Figure~\ref{fig:Audio_Interaction}, and its corresponding transcription in Table~\ref{tab:transcription_of_Interaction}, we can observe that this audio sample interleaves segments of short silences with segments of the teacher's speech, usually indicating that the lecturer is conversing with a~student.

\unskip

\begin{table}[H]
    \caption{Transcription of Figure~\ref{fig:Audio_Interaction}.}
    \label{tab:transcription_of_Interaction}
    \begin{tabular}{|p{13.35cm}|}
      \noalign{\hrule height 1pt}
        Context: The lecturer answers a question about an exercise\\
       \noalign{\hrule height 0.5pt}
        coming this way, which I'm going to call I. I plus I R one will equal I R two. Okay, that's going to hold true. But~I'm applying Ohm's law on resistance one. Okay? Yes? Come~on. \\
\noalign{\hrule height 1pt}
    \end{tabular}
\end{table}
    
\begin{figure}[H]
 \includegraphics[scale=0.37]{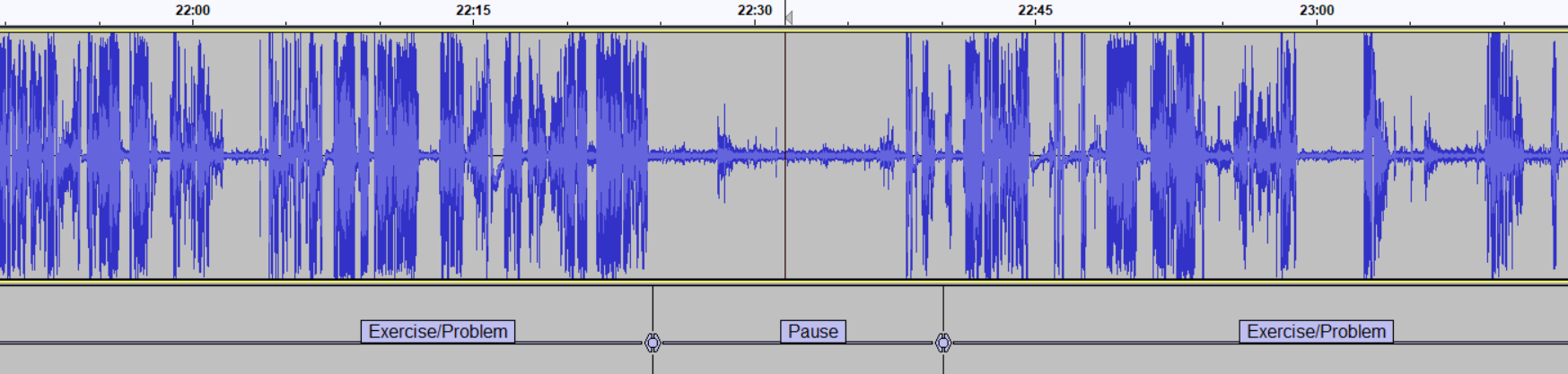}
    \caption{Audio sample of a Pause segment between Exercise~segments.}
    \label{fig:Audio_Exercise}
\end{figure}

Exercise: Figure~\ref{fig:Audio_Exercise} and its corresponding transcription in Table~\ref{tab:transcription_of_Exercise} shows an audio sample that also interleaves periods of silence with teacher's speech, like in Interaction, but~in this case the duration of the segments of speech and silence are generally longer than in Interaction. The~silences in Figure~\ref{fig:Audio_Exercise} mainly happen when the teacher is writing on the blackboard and stops sporadically in order to check if students are able to follow the explanation. Looking at the content of Table~\ref{tab:transcription_of_Exercise}, we can conclude that the lecturer is solving an Exercise due to the use of variables and formulae and~the fact that an equation for a specific electric circuit is being~solved. 

\unskip

\begin{table}[H]
    \caption{Transcription of Figure~\ref{fig:Audio_Exercise}.}
    \label{tab:transcription_of_Exercise}
    \begin{tabular}{|p{13.35cm}|}
      \noalign{\hrule height 1pt}
        Context: The lecturer is solving an exercise about the necessary values on an electrical circuit so that certain diodes conduct electricity or not\\
      \noalign{\hrule height 0.5pt}
        it's going to be off. OK? Because I've seen it before. Okay. So, for~what values of the diode is, diode one is on? So I will have to clear. I'm going to put greater than or equal to. I will clear from there. I'm going to clear E from this equation, to~see the values of E for which the diode conducts. Okay? Values of E for which diode one conducts. Let's look at diode two. In~this case it wouldn't be necessary, but~I'm going to analyse it so you can see. I don't know what it would take, because~I'm going to get a negative value. V of two. What does V of two equal? Look, where is the plus of the voltmeter? At A. And~where is the minus of the voltmeter? At C, which is D. Therefore, it will be V A minus V D. V A D and V A D is minus V of A.\\
        \noalign{\hrule height 1pt}
    \end{tabular}
\end{table}

\begin{figure}[H]
\includegraphics[scale=0.3]{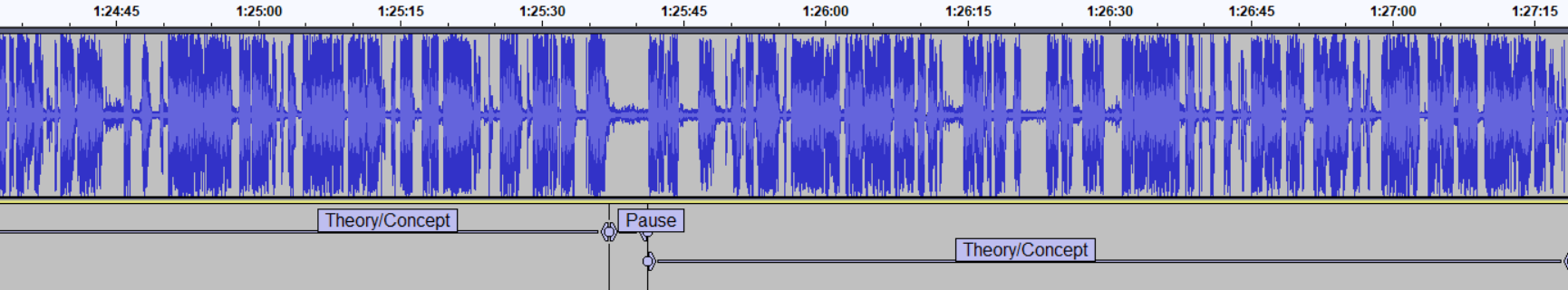}
    \caption{Audio sample of a Pause segment between Theory~segments.}
    \label{fig:Audio_Theory}
\end{figure}

Theory/Example/Organization/Digression/Other:  We grouped all these labels together because clearly and distinguishable audio features that discriminated among the teaching activities did not exist, as all of them were consistent with the audio signal of a monologue of the lecturer's speech. This is reflected in Figure~\ref{fig:Audio_Digression} (Digression), Figure~\ref{fig:Audio_Organization} (Organization), Figure~\ref{fig:Audio_Theory} (Theory), and~their corresponding transcriptions in Tables~\ref{tab:transcription_of_Digression}, \ref{tab:transcription_of_Organization} and \ref{tab:transcription_of_Theory}. We needed to distinguish them based on the content of the speech. In~Table~\ref{tab:transcription_of_Theory}, we can see that the lecturer is explaining a concept belonging to the syllabus of the course, specifically, the~main characteristics of the Zener diode (a special type of diode designed to reliably allow current to flow backwards), and~clarifies certain figures on the notes that usually confuse the~students.

\unskip

\begin{table}[H]
  \caption{Transcription of Figure~\ref{fig:Audio_Theory}.}
    \label{tab:transcription_of_Theory}
    \begin{tabular}{|p{13.35cm}|}
       \noalign{\hrule height 1pt}
        Context: The lecturer is explaining the main properties of the zener\\
       \noalign{\hrule height 0.5pt}
        In some books, even in some of the figures in the notes it says this and that confuses you, OK? It doesn't mean that this is the negative terminal and this is the positive terminal, but~when it conducts in reverse I have a source of V Z value with that polarity. OK? It's different always for the characteristic curve what we take as a reference is V D and D where the plus D V D is the plus, it's the anode and the minus is the cathode. OK? Don't get confused with this. This simply means that if I'm in reverse I've got a source there with that polarity V Z. OK? Well, the~main property of the zener is that it can conduct in direct. I have a non-zero current when V D is greater than V T H. The~diode will be on, direct, and~it behaves like a normal diode. Okay? What does it mean it behaves like a normal diode? That its equivalent model is a voltage source of the, with~the same polarity. Notice that this is the plus and this is the minus, plus, minus, with~the same polarity as the zener diode, right, and~the threshold voltage value. This is direct conduction, OK? From here to there it would be direct and when the voltage is less than zero it's what we call reverse bias. So zener diodes have the property that they can also conduct the current can be greater than zero, when they're reverse biased. Okay? This value here is minus V Z when the diode terminal voltage is smaller than minus V Z. OK? The datasheets give me the value of positive V Z but but as long as it's a zener and I have to know that this zener voltage is negative. OK? And therefore the conditions that the voltage at the diode terminals has to be less than minus V Z. Because~it's a negative T.\\
        \noalign{\hrule height 1pt}
    \end{tabular}  
\end{table}

From the above exposition, we can observe the distinctive audio signals of those activities that involve some kind of students engagement such as Interaction or Exercise. We were thus able to extract useful information from the audio recordings related to the speed of utterance, pitch of voice and pausing and phrasing that helped distinguish this type of activities from those that were categorized as a monologue-style of the lecturer. For~those activities that represent an extended speech of the teacher, we were able to extract distinguishable features from the transcribed notes. Hence, we expected that exploiting  together audio features and text features would ease the task of segmenting and classifying academic activities from class~recordings.

\section{Methodological~Design}
\label{methods}

In our approach, we started by preprocessing the automated transcription and the audio signal so as to obtain a rich representation from pretrained models. Then, we used that information to train our classification system. These two stages are discussed in detail in the next two~subsections.

\subsection{Feature~Extraction}

We split both our automated transcriptions and the audio files in frames of one second, so that for each frame we had the part of the transcription that was said in that second and the corresponding audio of that second. We chose one-second frames as we thought this offered a good balance between the granularity for the segmentation task and the computation~cost.

\begin{figure}[H]
\includegraphics[scale=0.27]{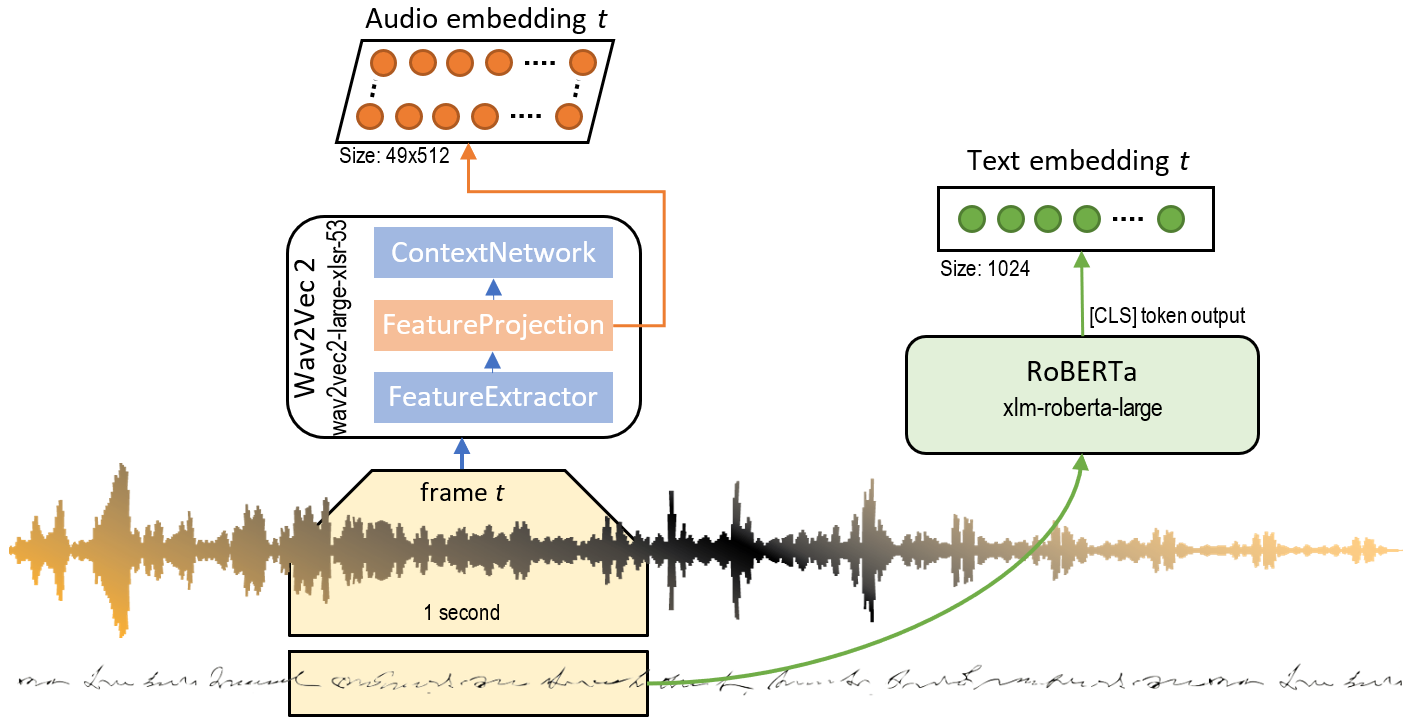}
    \caption{Extraction of text and audio~embeddings.}
    \label{fig:embeddings}
\end{figure}

Our system used XLM-RoBERTa~\cite{conneau2020unsupervised} to generate embeddings of the automated transcriptions. For~the audio signal, we used the Wav2Vec 2 feature extractor~\cite{Baevski2020wav2vec2A} to obtain the latent speech representation of the raw audio. {The scheme of this preprocessing stage can be observed in Figure~\ref{fig:embeddings}.} The next two subsections explain this stage in more~detail.

\unskip

\subsubsection{Text Feature~Extraction}

In order to work with the text coming from automated lecture transcriptions, we need to employ embeddings; i.e.,~a numerical representation of the text that captures as much of the linguistic meaning of said text as~possible. 

Embeddings are learned encodings that convert text data into numerical data. The~embeddings allow one to capture representative text features such as the meaning or use of a word and thereby words with similar meaning or words used in the same contexts end up having similar representations. Word embeddings are the most typical ones when working with text. We note that our task deals with text coming from automated transcriptions which may contain some minor errors and lack~punctuation.

We used XLM-RoBERTa to obtain embeddings from the transcription. XLM-RoBERTa is a multilingual language transformer-based model trained on 100 languages that offers rich text representations. The~transformer is a novel neural network architecture designed to work in sequence-to-sequence tasks with the capability to handle long dependencies with ease~\cite{VaswaniSPUJGKP17}. The~reason we used a multilingual model is because our intention is to extend our processing tool to work with other languages in the~future. 

Words that start in one frame and end in the next one are repeated, so such words are the last ones in one frame and the first ones in the subsequent frame. For~example, the~phrase ``Okay? Values of E for which diode one conducts.'' of Figure~\ref{tab:transcription_of_Exercise}  would be split this way: Okay? | Values of E | E for which | which diode one | one conducts.

The words of each frame were fed to XLM-RoBERTa and the output corresponding to the [CLS] 
token on the last layer was used as an embedding. The~[CLS] token is a special token that is used for sentence-level classification. This token serves as a sort of sentence embedding as it encodes all the words in the input of XLM-RoBERTa in a single embedding. This way, all the words contained in a one-second frame are represented by one embedding of 1024 values. Finally, we stored all the embeddings in files for later~use.

\subsubsection{Audio Feature~Extraction}

We used the  Wav2Vec 2 multilayer convolutional feature extractor to obtain latent speech representations from the raw input audio. Wav2Vec 2 is a transformer-based model that uses a self-supervised approach to learn representations from raw audio data. This model first encodes the speech audio via a multilayer convolutional neural network, obtaining latent speech representation that are later fed to a transformer network in order to build contextualized~representations.

For our task, we used the Wav2Vec 2 feature extractor, which consists of seven consecutive one-dimensional convolutions with 512 channels and respective kernel sizes of (10, 3, 3, 3, 3, 2, 2) and stride of (5, 2, 2, 2, 2, 2, 2). For~a more detailed technical description, we refer the interested reader to the original paper~\cite{Baevski2020wav2vec2A}. We split the raw audio signal in frames of one second, and~each of these frames was given as input to the feature extractor. We then obtained the embeddings from the extractor and saved them on files for later use. Each one-second frame became an embedding of size (49, 512).

\subsection{Classifier~Architecture}
\label{architecture}

\begin{figure}[H]
   \includegraphics[scale=1.05]{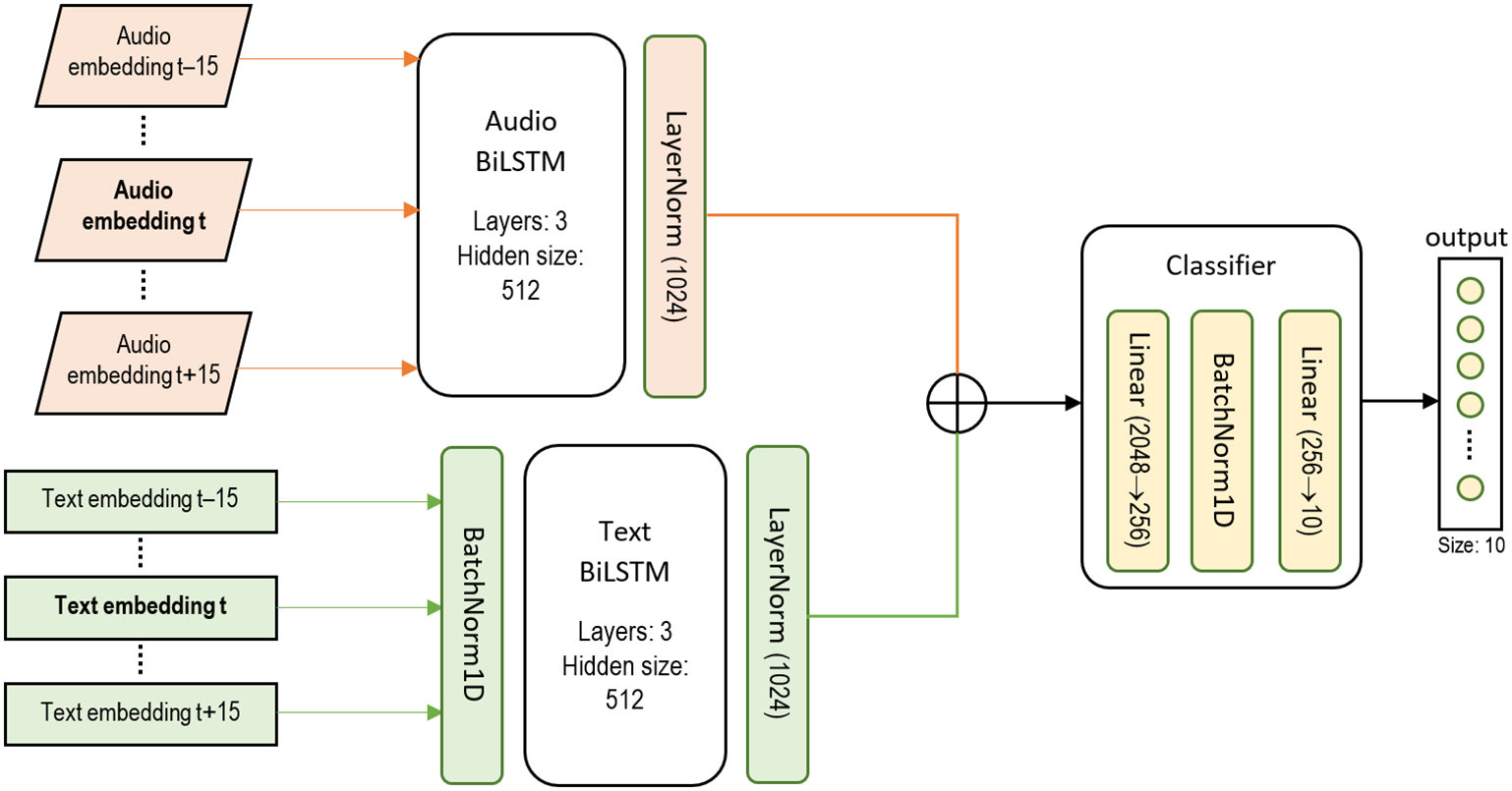}
    \caption{Diagram of our classifier~model.}
    \label{fig:model}
\end{figure}

The architecture of our model is composed of two bidirectional LSTM (BiLSTM) layers of 512 units, one for processing the audio signals and one for processing the text transcriptions, and~the classifier. {A simple diagram of our model can be seen in Figure~\ref{fig:model}}. The~classifier is a fully connected layer of 2048 units with a Gelu activation followed by the output layer with a Softmax~activation.

The model of the processing tool receives as input a sequence of frames, constructed as $\{f\textsubscript{t$-$N}...f\textsubscript{t$-$1}, f\textsubscript{t}, f\textsubscript{t+1}...f\textsubscript{t+N}\}$, where $f\textsubscript{t}$ is the frame we want to classify, and~we add the previous and posterior N frames as additional information. Each frame is composed by its audio embedding and its text embedding. The~text and audio embeddings are divided, and~the text embeddings are normalized. The~audio embeddings have already been normalized by the Wav2Vec 2 feature~encoder.

We forward each embedding to its corresponding BiLSTM, normalize the outputs and concatenate them. Then, we forward the result to the classifier. We then normalize the classifier's output of the fully connected layer after the Gelu activation and~forward it to the output layer. The~output layer returns the probability that $f$\textsubscript{t} belongs to each class (teaching activities of Figure~\ref{fig:LabelHierarchy}).

We used BiLSTMs to make the most of the context of a frame, i.e.,~the $N$ previous and $N$ posterior frames. We chose to simply concatenate the output of both BiLSTMs layers so that the model was able to learn to identify the activities based on the transcription and audio information~simultaneously.

We tried different values of $N$, the~number of previous and posterior frames that act as additional information. We started with 120 and~tried to reduce it without compromising the performance of the model. We observed that for $N=15$, the~results were slightly worse, but~the reduction in use of memory and computational cost was worthwhile.  The~initial loss was plotted over a wide range of learning rates and the final selected value for the initial learning rate was $0.001$.
We used Adam as the optimizer, and~employed a one-cycle learning rate policy~\cite{Smith2019SuperconvergenceVF}.

\section{Results}
\label{results}

This section presents the results of our work on the recognition of teaching activities of a lecturer from university class~recordings.

In the recordings of the lectures registered at our university prior to 2020, only the lecturer's voice was available. As~commented before, our dataset was composed of 34 lecture recordings of different courses, such as mathematics, electronic devices, physical oceanography, statistics, etc. Out of these 34 recordings, the~transcriptions of 14 of them were manually revised. In~total, the~recordings were approximately 3700 min long, of~which 1600~min were lectures given by women and 2100 min by~men. 

\begin{figure}[H]
  \includegraphics[scale=0.5]{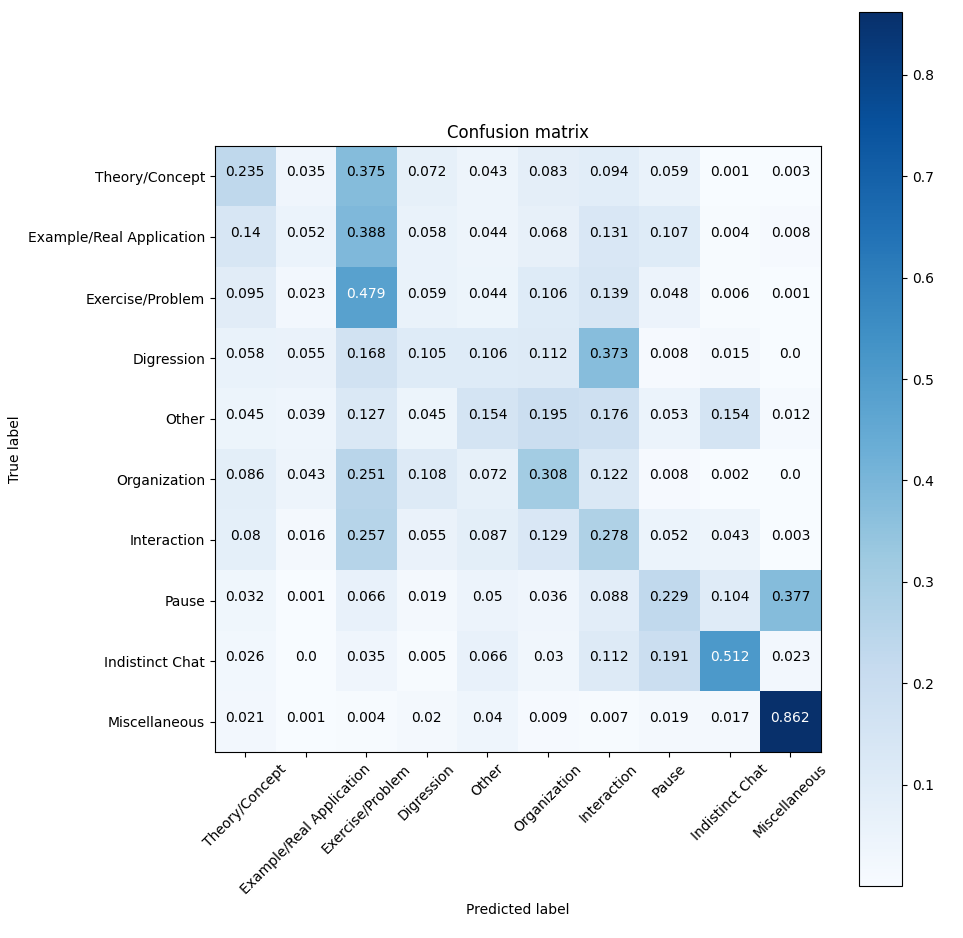}
    \caption{Confusion~matrix.}
    \label{fig:Confusion_Matrix}
\end{figure}

The model was evaluated using a partition of 90\% for training and 10\% for evaluation. We shuffled the dataset and stratified the partition, so the evaluation held approximately one tenth of the data of each class or teaching activity. As~the dataset was imbalanced, a~class weight tensor to account for this imbalance was computed. We report the values of precision, recall and F-score for each class in Table~\ref{tab:metrics} and the confusion matrix of the results in Figure~\ref{fig:Confusion_Matrix}.

In the confusion matrix of Figure~\ref{fig:Confusion_Matrix}, rows show the true label of the segments, i.e.,~the label we manually assigned to the segments, and~columns represent the class predicted by our model. The~values on the diagonal are the number of true positives (TP); for each class, the~values in the columns show the false positives (FP) and the values in the rows show the number of false negatives (FN).

The best performing class is Miscellaneous, with~an F-score of 0.875, followed by Indistinct Chat, Exercise/Problem and Interaction with F-scores 0.437, 0.391 and 0.367, respectively. In~the confusion matrix, the~low values of the last row indicate that Miscellaneous is an easily distinguishable class, as~our model correctly classifies the majority of Miscellaneous frames. However, a~considerable amount of Pause frames are classified as Miscellaneous, as~shown by the value 0.377 in the last column of Figure~\ref{fig:Confusion_Matrix}.

\begin{table}[H]
  \caption{Precision, recall and F-score by~class.}
    \label{tab:metrics}
    \setlength{\tabcolsep}{21.3pt}
    \begin{tabular}{lrrr}
        \toprule
        Label & Precision & Recall & F-Score \\ \midrule
        Theory/Concept & 0.279 & 0.235 & 0.255\\ 
        Exercise/Problem & 0.330 & 0.479 & 0.391\\
        Example/Real Application & 0.169 & 0.052 & 0.079\\
        Organization & 0.195 & 0.308 & 0.238\\
        Interaction & 0.538 & 0.279 & 0.367\\
        Digression & 0.037 & 0.105 & 0.055\\
        Other & 0.013 & 0.154 & 0.024\\
        Indistinct Chat & 0.381 & 0.512 & 0.437\\
        Pause & 0.254 & 0.229 & 0.241\\
        Miscellaneous & 0.889 & 0.862 & 0.875\\\midrule 
    \end{tabular}  
\end{table}

Moreover, the~results obtained for the Theory/Concept and Organization classes indicate that, although~the values are worse than for the other classes, the~model is able to distinguish Theory/Concept and Organization frames. Specifically, the~Organization class has a distinctive vocabulary (dates, grading system, submitting exercises) that differentiates it from the other classes. Conversely, the~metrics of the classes Digression, Other and Example/Real Application fall behind the rest of the classes. The~reason for the poor performance of Digression and Other can be found in the low number of samples of these two classes in the dataset, as~these activities occurred infrequently during the lectures, and~their duration was usually rather short. In~the fourth and fifth column of Figure~\ref{fig:Confusion_Matrix}, we can observe that our model hardly predicts frames as belonging to classes Digression and Other. Furthermore, frames of Digression are predicted as belonging to Interaction, while Other frames are misclassified as Organization, Interaction and Indistinct Chat. Furthermore, in~the third column, it should be noted that our model is biased towards Exercise/Problem, and~predicts a considerable proportion of Theory/Concept, Example/Real Application, Organization and Interaction as belonging to Exercise/Problem.

Taking into account these results, it is worth noting that the best performing classes (Miscellaneous, Indistinct Chat, Exercise/Problem and Interaction), together with Pause, are the classes that were more readily distinguishable by means of the audio signal. This seems to indicate that the model is paying more attention to audio features than text features, as~can be observed by the low value of the errors in the confusion matrix, particularly in the rows and columns of Miscellaneous, Indistinct Chat and~Pause.

A possible way to address the audio-skewed behavior of our model would be to modify the way in which the audio and text features are forwarded to the classifier. Another line worth exploring would be to divide the model into two classifiers and classify the audio-dependent~activities first.

\section{Discussion}
\label{discussion}

In this section, we attempt to put into perspective the results of our approach. While comparing to other multimodal systems is not easy because each one addresses a different task with different technologies, we can say that the best accuracy results obtained with multimodal classification across various areas is around 80\%. For~example, the~best predicted class in the emotion classification of the work by Yoon et al. (2018) was the class ``happy'' with an accuracy of 79.08\%, and~the speech six-intention classification model achieved a 83.10\% average accuracy~\cite{GuChen17}. Better results around 93\% were achieved in music genre classification when using audio, text and images~\cite{OramasBNS18}. As~we commented in Section~\ref{multimodal}, we must also stress that these systems work in limited application contexts and, more importantly, they use datasets that include professional transcription services~\cite{Yoon18} or manually transcribed text~\cite{GuChen17}.

The performance of our model was below the best accuracy achieved by the aforementioned approaches as we faced several limitations such as the use of a small dataset that contained a low number of samples (34 lesson recordings), potential labeling errors as well as the errors inherent in automated~transcriptions.

On the other hand, we also offer some advantages over other models:

\begin{itemize}
    \item Our proposal is independent of the audio recording system and does not need proprietary audio recorders such as in LENA~\cite{WangPMC14}.
    \item Our model is independent of the automated transcription tool and can be used with the transcriptions provided by any service such as YouTube, Apple's Siri, Zoom video communications or others. In~our case, we used the open-source MLLP transcription and translation platform~\cite{Martinez-VillarongaAAJ13,MiroSCTJ15}, which turned out to translate scientific and technical terms better than YouTube. 
    \item The use of transformer-based models broadens the applicability of the classification architecture to different thematic contexts and different languages. Our model is thus applicable to a large variety of subjects of different natures and is extensible to other~languages.
\end{itemize}

\section{Conclusions and Future~Work}
\label{conclusions}

The classification model presented in this paper represents the first step toward a mechanism that enables students to find specific contents in an audio recording. In~a nutshell, the~output of the model can be viewed as the ``table of contents'' of a lesson given by a lecturer wherein the different teaching activities along the recording are index-time-stamped. We identified a set of activities that characterized the spoken academic discourse and we designed a classifier using a transformer-based language model, specifically a version of the BERT family transformer models, that exploited both audio and text features. The classifier architecture was based on two LSTM neural networks to process the audio signals and the text transcriptions. The~results showed that some teaching activities were better identifiable with the audio signal while others required resorting to the text transcriptions as the main data source. Overall, the~promising obtained results open up interesting ways of improvement and~challenges.

{As for future work, we identify two lines of action. In~the line of \textbf{technical improvements}, we aim to improve the accuracy of our model by testing different mechanisms. With~the purpose of addressing the lack of attention of the model to the text features, we propose to address the classifier model hierarchically: first, we will use spectrograms of the audio segments to distinguish between silence, noise and~student talk from the teacher's speech. Then, we will classify the teacher's talk into the different types of activity according to the transcriptions and the context. We think that this new text classifier will output better results using a more intelligent segmentation of the input data, splitting the transcripts according to the small pauses in the speech detected in the audio.} Another technique for increasing the accuracy is to fine-tune XLM-RoBERTa with the automated transcriptions on the repository. A~further refinement could also be training our own language model tailored for the spoken academic~discourse.

Regarding the \textbf{scope of the work}, our intention is to improve the learning experience of the student by facilitating the human-recording interaction as well as to boost and enhance learning from video class recordings, which is the primary learning resource in university classes. {To that end, our multimodal classification algorithm of teaching activities will be integrated into a tool for both students and academic staff. It will make it easy for \textbf{students} to view long class recordings, providing direct access to specific contents. Ultimately, we aim to develop a web application tool that enables students to select the desired type of teaching activity so that the playback jumps straightforwardly to the desired point. It will also aid \textbf{teachers} in identifying the type of activity they are doing during a lesson as well as retrieving valuable information about how long the teacher devotes to explaining the theory and solving exercises, or~how many times the teacher interacts with students and attempts to engage them in the classroom. These data can then be eventually used to cross-check the results of the teaching evaluation questionnaires and study correlations between the use of teaching activities and student satisfaction. This will allow them to enhance their teaching style and discuss how teaching and learning might be improved in the class.}

\vspace{6pt}

\authorcontributions{Conceptualization, O.S. and E.O.; methodology, O.S.; software, O.S.; validation, O.S. and E.O.; formal analysis, E.O.; investigation, O.S. and E.O.; resources, O.S. and E.O.; data curation, O.S.; writing---original draft preparation, O.S.; writing---review and editing, E.O.; supervision, E.O.; funding acquisition, E.O. All authors have read and agreed to the published version of the manuscript. 
}

\funding{This research was funded by the project CAR: Classroom Activity Recognition of GENERALITAT VALENCIANA. CONSELLERÍA D'EDUCACIÓ grant number PROMETEO/2019/111.}

\conflictsofinterest{The authors declare no conflict of~interest.}




\begin{adjustwidth}{-\extralength}{0cm}

\reftitle{References}



\end{adjustwidth}
\end{document}